\documentclass[11pt]{article}

\usepackage{acl}

\usepackage{times}
\usepackage{latexsym}
\usepackage[T1]{fontenc}
\usepackage[utf8]{inputenc}
\usepackage{microtype}
\usepackage{inconsolata}
\usepackage{graphicx}
\usepackage{amsmath}
\usepackage{amssymb}
\usepackage{booktabs}
\usepackage{multirow}
\usepackage{subcaption}

\newcommand{\darts}{\texttt{DARTS}}
\newcommand{\R}{\mathbb{R}}
\DeclareMathOperator{\ReLU}{ReLU}
\DeclareMathOperator{\softmax}{softmax}

\title{\raisebox{-0.5ex}{\includegraphics[height=1.2em]{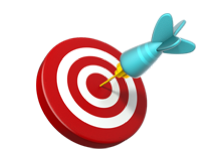}} \darts: Decoder-Aware Representation Tuning via \\ Surgery for Model Merging}

\author{
	\textbf{Aaryan Ajay Sharma}$^{1,2,*,\dagger}$ \quad \textbf{Sai Nishanth Padala}$^{1,*,\dagger}$ \quad \textbf{Seganrasan Subramanian}$^{1}$ \\
	$^1$ServiceNow \quad $^2$University of Twente \\
	\texttt{aaryan.sharma@utwente.nl}, \texttt{p.snishanth996@gmail.com}, \\
	\texttt{seganrasan.subramanian@servicenow.com}
}

\begin{document}

\maketitle

\begin{abstract}
Model merging combines multiple task-specific fine-tuned LLMs into a single multi-task model without additional training. However, merged models are known to suffer from \emph{representation bias}: systematic drift between the merged model's hidden states and those of each individual source model. Prior work~\citep{yang2024representation} study and mitigate this bias for encoder-based vision models using a lightweight correction module trained with L1 loss. However, such bias is not studied for decoder models due to their autoregressive nature. Therefore, we first analyze the problem of representation bias in decoder models, and show two challenges absent in encoders: (1) the causal attention mask causes bias to \textbf{accumulate} across token positions, requiring position-dependent correction; and (2) not all token positions are equally important, i.e., high-entropy (decision-critical) positions matter far more than low-entropy ones. To address these challenges, we propose \underline{\textbf{D}}ecoder-\underline{\textbf{A}}ware \underline{\textbf{R}}epresentation \underline{\textbf{T}}uning via \underline{\textbf{S}}urgery~(\darts). \darts\ employs a novel {entropy-weighted L1 loss} to upweight correction at high-entropy positions where errors most affect generation quality, and a per-position additive bias that captures position-dependent error without overparameterization. We perform extensive evaluation on three domains: code generation (HumanEval), mathematical reasoning (GSM8K), and instruction following (AlpacaEval) on  Llama-2-7B models, and show \darts\ achieves significant improvement over the standard surgery approach while adding negligible parameters (0.1\% of total parameters).
\end{abstract}

\section{Introduction}
\label{sec:introduction}

\begin{figure}[t]
  \centering
  \includegraphics[width=\columnwidth]{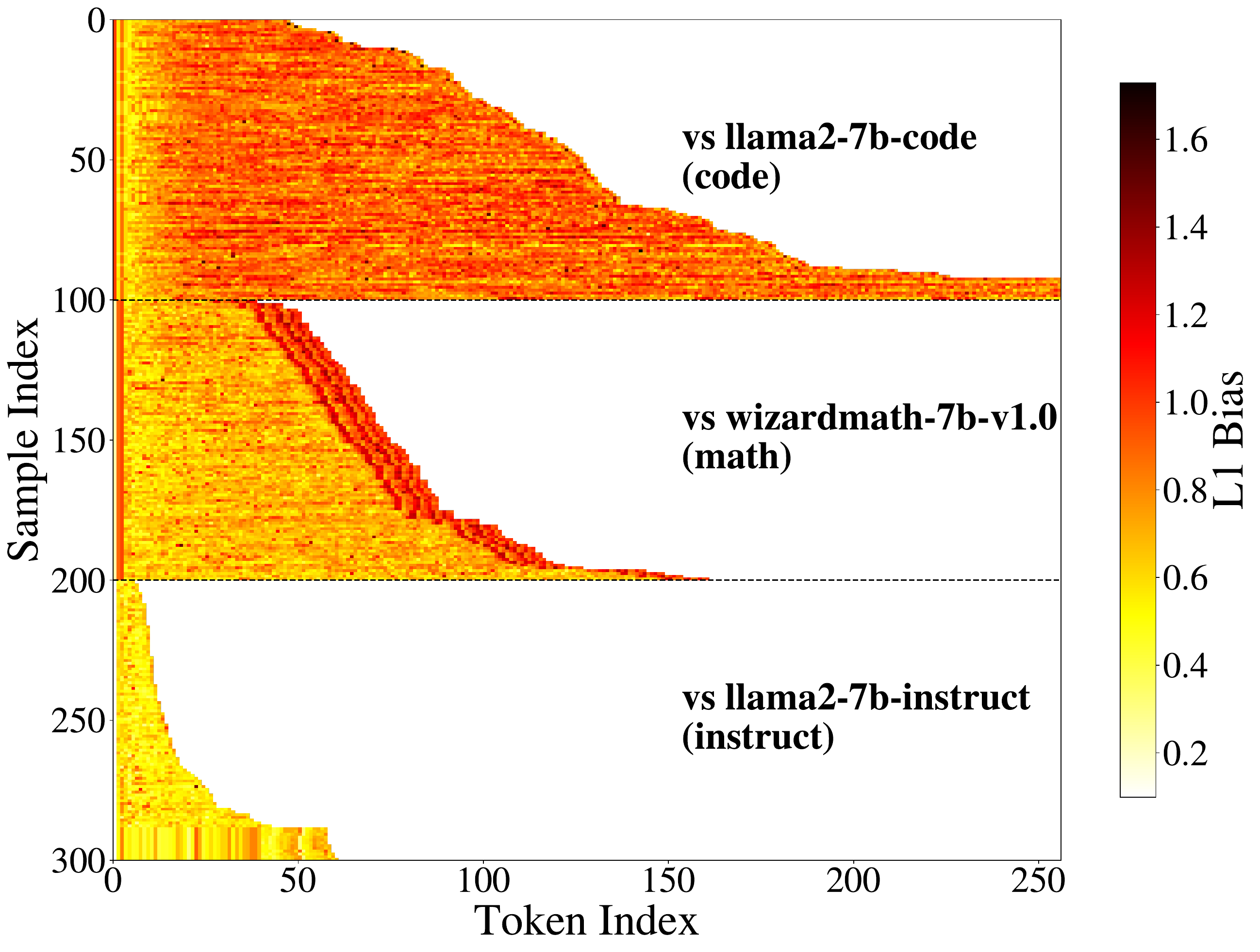}
  \caption{Per-token L1 representation bias between a merged Llama-2-7B model and each individual fine-tune, measured at the final transformer layer. Each row is a calibration sample; each column is a token position. Across all three domains (code, math, instruct), we see an increasing trend of bias with token position due to error accumulation under the causal attention mask. This position-dependent structure motivates the decoder-aware corrections proposed in \darts.}
  \label{fig:teaser}
\end{figure}

Large language models (LLMs) have demonstrated remarkable capabilities across a wide range of tasks, from code generation~\citep{chen2021humaneval,roziere2023code} and mathematical reasoning~\citep{cobbe2021gsm8k,luo2023wizardmath,wei2022chain} to open-ended instruction following~\citep{ouyang2022training}. However, these capabilities are typically obtained through task-specific fine-tuning~\citep{hu2021lora,touvron2023llama}, producing a proliferation of specialized models, each of which excels on its target domain but requires independent deployment. As the number of specialized models grows, so does the cost of serving them, making multi-task consolidation increasingly attractive.
 
  \begingroup
  \renewcommand\thefootnote{}%
  \footnotetext{%
    \textsuperscript{*}Equal contribution. Author order determined by \href{http://qrng.anu.edu.au/random-binary/}{quantum random number generator} over Google Meet.\\
    \textsuperscript{$\dagger$}Corresponding authors.\\
    Work done during Aaryan's internship at ServiceNow.%
  }%
  \addtocounter{footnote}{-1}%
  \endgroup
  
\emph{Model merging}~\citep{wortsman2022modelsoups,ilharco2023editing,yadav2023tiesmerging,yang2024adamerging,yadav2025a} offers an elegant solution to this problem: combine the weights of multiple fine-tuned models into a single model that inherits capabilities from all source models, \textit{without any additional training data or computation}. Methods such as Task Arithmetic~\citep{ilharco2023editing}, TIES-Merging~\citep{yadav2023tiesmerging}, and DARE~\citep{yu2024language} have shown that simple arithmetic operations in weight space (such as adding scaled task vectors, resolving sign conflicts, or dropping redundant parameters) can produce surprisingly effective multi-task models. This has made model merging a standard tool in the open-source LLM ecosystem~\citep{goddard2024arcee}, where practitioners routinely merge instruction-tuned, code-specialized, and math-specialized models into unified checkpoints.

Despite this progress, a substantial performance gap persists between merged models and their individual fine-tuned counterparts. Merged models frequently degrade on one or more constituent tasks~\citep{ilharco2023editing,yadav2023tiesmerging,yang2024adamerging}, and the severity of this degradation varies unpredictably across tasks, merging methods, and model scales. Understanding \emph{why} merging hurts performance, and finding principled ways to close the gap, remains an open challenge.

\citet{yang2024representation} recently offered a compelling explanation for this degradation. By visualizing the hidden representations of merged vision models (ViT-B/32, ViT-B/16, ViT-L/14 on eight classification tasks), they demonstrated that merged models suffer from \textbf{representation bias}: a systematic discrepancy between the feature representations extracted by the merged model and those of each individual fine-tuned model. Crucially, they showed that representation bias \emph{correlates} with task performance: merging methods that produce smaller representation bias (e.g., AdaMerging~\citep{yang2024adamerging}) consistently outperform those with larger bias (e.g., Weight Averaging). Building on this insight, they proposed \textit{representation surgery}: a lightweight low-rank module trained with L1 loss to predict and subtract the representation bias, achieving significant performance gains across all merging methods.

However, the original surgery framework was designed exclusively for \textbf{encoder-based vision models}. In ViT architectures~\citep{dosovitskiy2021vit}, bidirectional self-attention ensures that each token (image patch) is contextualized by all other tokens equally, producing representation bias that is roughly \emph{uniform} across token positions. A position-agnostic surgery module, which applies the same correction at every position, is well-suited to this setting. Likewise, the uniform L1 loss used for training treats all token positions as equally important, which is reasonable when every patch contributes similarly to the classification decision.

\textbf{Decoder-based LLMs are fundamentally different.} The autoregressive transformer architecture~\citep{vaswani2017attention,brown2020language} processes tokens left-to-right under a causal attention mask: token $t$ can only attend to tokens $0, 1, \ldots, t{-}1$. This architectural constraint has two consequences for representation bias that render the original surgery framework inadequate:

\textbf{First, representation bias in decoder models is position-dependent.} Because each token's representation is computed from a strictly causal context, any bias introduced by weight-space merging at earlier positions propagates forward through the attention mechanism. Later tokens aggregate not only the original merging error at their own position but also the accumulated errors from all preceding positions. \figurename~\ref{fig:teaser} empirically shows this effect: plotting per-token L1 distance between merged and individual model representations reveals a clear \emph{increasing} trend, with later positions exhibiting 2--3$\times$ higher bias than early positions. A position-agnostic surgery module, which applies identical corrections at every position, is structurally incapable of capturing this pattern, forcing it to compromise between under-correcting later tokens and over-correcting earlier ones.

\begin{figure*}
    \centering
    \includegraphics[width=1\linewidth]{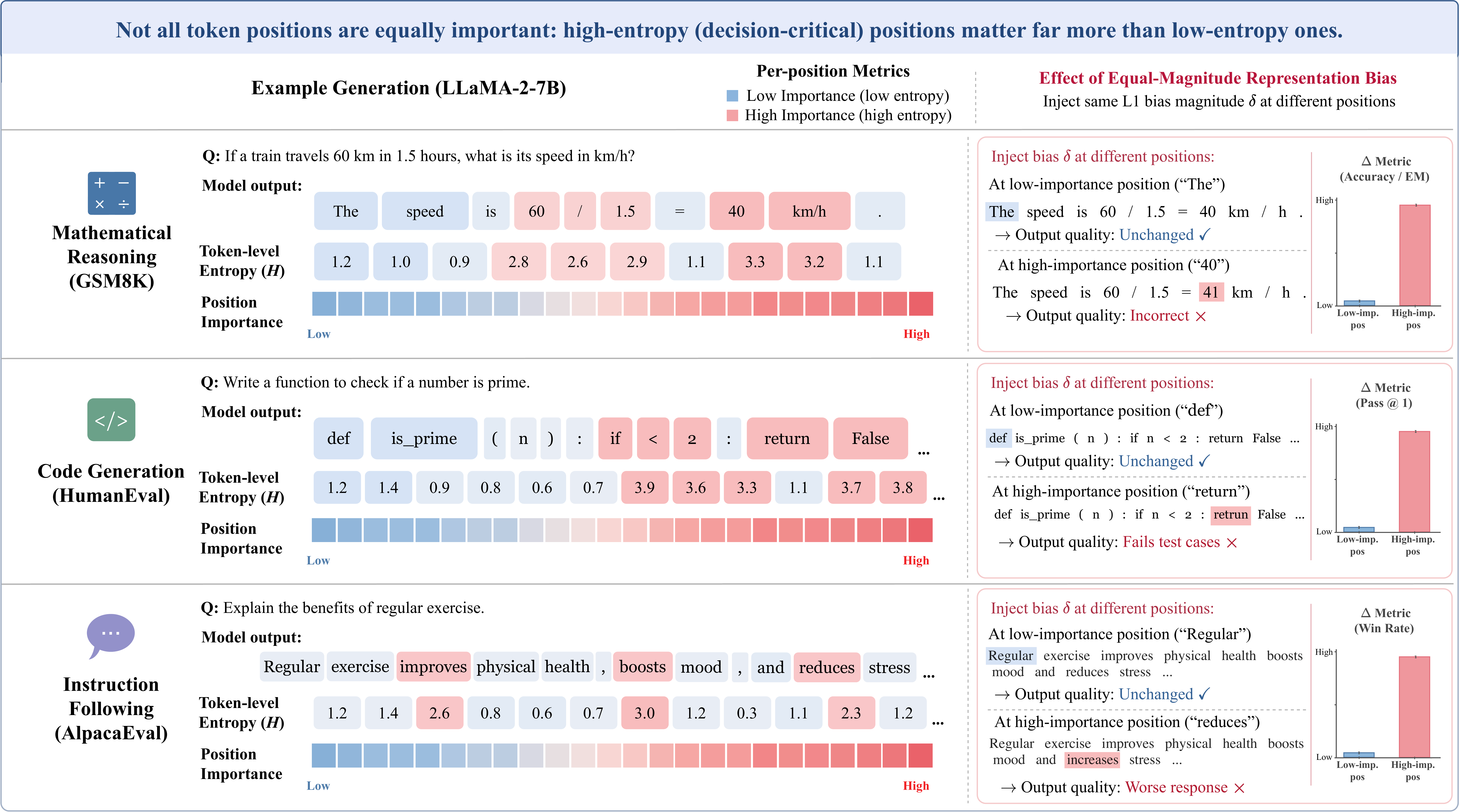}
    \caption{{Non-uniform importance of token positions in decoder models. We show three representative domains (math, code, instruction).
For each example, we illustrate token-level entropy ($H$) and visualize position importance (higher entropy = higher importance). Injecting the
same-magnitude representation bias $\delta$ at low-importance positions has negligible impact, while injecting at high-importance positions causes significant performance degradation. This motivates entropy-weighted losses that focus corrections where they matter most.}}
    \label{fig:elw1_motivation}
\end{figure*}

\textbf{Second, not all token positions are equally important for generation quality.} In autoregressive LLMs, the model's prediction at each position determines the next token in the generated sequence. At some positions, the model is highly confident and assigns most probability mass to a single token (e.g., closing brackets, deterministic syntax, common phrases). At other positions, the model distributes probability across many plausible continuations; these are the \emph{decision-critical} positions where the choice of next token meaningfully affects the quality and correctness of the generated output. Small hidden-state errors at high-entropy positions can flip the argmax prediction and cascade through subsequent tokens, while comparable errors at low-entropy positions are absorbed without consequence. The standard $L_1$ loss~\citep{yang2024representation} treats all positions uniformly, wasting the surgery module's limited correction capacity on positions that do not affect downstream generation. \figurename~\ref{fig:elw1_motivation} illustrates this idea.

To address these decoder-specific challenges, we propose \underline{\textbf{D}}ecoder-\underline{\textbf{A}}ware \underline{\textbf{R}}epresentation \underline{\textbf{T}}uning via \underline{\textbf{S}}urgery~(\darts), which adapts representation surgery to the autoregressive setting through two complementary innovations. First, we introduce \textbf{entropy-weighted $L_1$ loss (EWL1)}, which weights the per-position correction penalty by the target model's prediction entropy, focusing the surgery module's capacity on decision-critical positions where errors most affect generation quality. Second, we augment the low-rank content correction with a \textbf{position correction module}, a direct per-position additive bias table that learns independent correction vectors for each token position, capturing the structured position-dependent bias without network overparameterization. We evaluate \darts\ on Llama-2-7B~\citep{touvron2023llama} models merged from three specialized fine-tunes (code generation, mathematical reasoning, and instruction following), showing that it achieves significant improvement over the original surgery approach.

The main contributions of this paper are as follows:
\begin{enumerate}
    \item We analyze the problem of representation bias in decoder models and empirically demonstrate two challenges absent in encoder models: \textbf{position-dependent bias accumulation} under the causal attention mask, and \textbf{non-uniform position importance} arising from the autoregressive generation process.
    \item To address these challenges, we propose \darts, which uses a novel \textbf{entropy-weighted $L_1$ loss} and a \textbf{position correction module}, to upweight correction at decision-critical positions, and perform position-dependent bias corrections.
    \item We perform extensive experiments to show that \darts\ reduces causal representation bias and significantly improves MTL performance on a variety of different tasks while adding merely 0.1\% of parameters to the model.
\end{enumerate}

\section{Related Work}
\label{sec:related-work}

\paragraph{Model Merging for Multi-Task Learning.}
Model merging has emerged as a practical alternative to traditional multi-task learning~\citep{crawshaw2020mtl,zhang2021mtlsurvey}, which requires centralized data collection and joint training. Early work showed that simply averaging model weights can improve generalization for models trained on the \emph{same} task~\citep{wortsman2022modelsoups}, and subsequent methods extended this idea to merging models trained on \emph{different} tasks. Task Arithmetic~\citep{ilharco2023editing} introduced the concept of \emph{task vectors}, defined as the difference $\tau_i = \theta_i - \theta_{\text{pre}}$ between fine-tuned and pretrained weights, and showed that adding scaled task vectors to the pretrained model produces effective multi-task models. TIES-Merging~\citep{yadav2023tiesmerging} addressed parameter sign conflicts and redundancy by trimming, electing signs, and merging only aligned components. DARE~\citep{yu2024language} randomly drops delta parameters and rescales survivors to reduce interference. Fisher-Merging~\citep{fisher2022merging} uses Fisher information to weight parameter importance, while RegMean~\citep{jin2023regmean} reweights based on training statistics. More recently, AdaMerging~\citep{yang2024adamerging} learns task-wise or layer-wise merging coefficients, Concrete~\citep{tang2026concretesubspacelearningbased} finds shared subspaces, and ZipIt~\citep{stoica2024zipit} merges models by aligning intermediate features. A comprehensive survey of these methods can be found in \citet{yadav2025a}. All of the above methods operate exclusively in weight space and do not explicitly address the resulting representation-level discrepancies.

\paragraph{Representation Surgery.}
\citet{yang2024representation} identified representation bias as a key obstacle in model merging: the hidden representations of merged ViT models~\citep{dosovitskiy2021vit,radford2021clip} diverge systematically from those of individual fine-tuned models, and this divergence correlates with task performance degradation. They proposed a lightweight adapter-like~\citep{houlsby2019adapter} correction module $\Phi(Z) = W_{\text{up}} \cdot \ReLU(W_{\text{down}} \cdot Z)$ trained with $L_1$ loss to subtract the bias from merged representations. This approach was validated on eight image classification tasks across three ViT architectures, achieving consistent gains when applied on top of Weight Averaging, Task Arithmetic, TIES-Merging, and AdaMerging. However, the original framework assumes position-agnostic correction and uniform loss weighting. These assumptions are well-motivated for bidirectional encoder models but, as we demonstrate, inadequate for autoregressive decoder models.

\paragraph{Inference-Time Interventions and Routing.}
Beyond static weight-space merging, recent work has explored inference-time methods for LLM capability consolidation. Activation steering~\citep{turner2023steering,li2023inference} dynamically injects or subtracts steering vectors during the generation loop to modulate model behavior at test time. Mixture-of-Experts (MoE) architectures~\citep{shazeer2017outrageously,fedus2022switch} and dynamic routing mechanisms~\citep{muqeeth2023soft} maintain multiple expert subnetworks and employ learned routers to select among them on a per-token or per-layer basis. Expert composition methods~\citep{tang2024merging} combine multiple specialized adapters or modules at inference time. While these approaches offer flexible capability control, they differ fundamentally from \darts\ in several practical dimensions. First, routing and MoE methods require loading multiple model blocks or expanded parameter sets in VRAM, whereas \darts\ maintains a single model footprint with only $\approx$0.1\% additional parameters. Second, inference-time steering and routing add dynamic computation branches that increase serving latency, while \darts\ applies a static correction with latency identical to the base model. Third, \darts\ produces a standard single-checkpoint model that serves as a drop-in replacement, requiring no specialized inference engine. We view these approaches as complementary: \darts\ is best suited for low-overhead, permanent consolidation of merged models, while inference-time methods offer greater flexibility at the cost of deployment complexity.

\section{Causal Representation Bias in Model Merging}
\label{sec:method}

We begin by establishing notation and defining the problem (Section~\ref{sec:preliminaries}), then empirically characterize the position-dependent structure of representation bias in decoder models (Section~\ref{sec:position-bias}). We then present the two core components of \darts: the position correction module (Section~\ref{sec:posdirect}) and entropy-weighted $L_1$ loss (Section~\ref{sec:ewl1}).

\subsection{Preliminaries}
\label{sec:preliminaries}

\subsubsection{Notation}

We consider $K$ task-specific models obtained by fine-tuning a shared pretrained model on tasks $\mathcal{T}_1, \dots, \mathcal{T}_K$. Let $M$ denote the merged model and $M^*_k$ denote the $k$-th individual fine-tune (the per-task ``target''). For an input sequence $x_{1:T}$ tokenized to length $T$, we denote:
\begin{itemize}
    \item $h^M_t \in \R^H$: the merged model's final-layer hidden state at position $t$
    \item $h^{*k}_t \in \R^H$: target model $k$'s corresponding hidden state
    \item $W_U^{*k} \in \R^{V \times H}$: target model $k$'s LM head (unembedding matrix)
\end{itemize}
with $H = 4096$ (hidden size, Llama-2-7B), $V = 32{,}000$ (vocabulary size), and $T_{\max} = 512$ (calibration max sequence length). For each task $k$, we learn a surgery module $\mathcal{S}_k$ that produces a corrected hidden state $\hat{h}_t = \mathcal{S}_k(h^M)_t$. We drop the $k$ subscript where clear from context.

\subsubsection{Task Vectors and Merging} Given a pretrained model $\theta_{\text{pre}}$ and a model fine-tuned on task $k$, $\theta_k$, the task vector is $\tau_k = \theta_k - \theta_{\text{pre}}$. Task Arithmetic~\citep{ilharco2023editing} merges $K$ tasks via:
\begin{equation}
\theta_{\text{merged}} = \theta_{\text{pre}} + \lambda \sum_{k=1}^{K} \tau_k
\label{eq:task-arithmetic}
\end{equation}
where $\lambda$ is a scaling coefficient. Other merging methods (TIES-Merging, DARE, AdaMerging) modify how task vectors are combined but share the same general framework~\citep{yadav2025a}.

\subsubsection{Representation Bias} Following~\citet{yang2024representation}, we define the per-position representation bias as the normalized $L_1$ distance between the merged and target model's hidden states:
\begin{equation}
\delta_t = \frac{1}{H}\|h^M_t - h^{*k}_t\|_1
\label{eq:rep-bias}
\end{equation}
In encoder models, \citeauthor{yang2024representation} showed that this bias is approximately uniform across token positions and correlates with downstream task performance. We extend this analysis to decoder models in the next section.


\subsection{Standard Surgery (Baseline)}
\label{sec:standard-surgery}

The original representation surgery~\citep{yang2024representation} employs a low-rank bottleneck with rank $r$ and content correction function $c(h) = W_{\text{up}}\,\ReLU(W_{\text{down}}\,h)$:
\begin{equation}
\hat{h}_t = h^M_t - c(h^M_t)
\label{eq:standard}
\end{equation}
where $W_{\text{down}} \in \R^{r \times H}$ and $W_{\text{up}} \in \R^{H \times r}$. Following~\citet{yang2024representation}, $W_{\text{up}}$ is zero-initialized so that the surgery module starts as the identity map, meaning an untrained module produces no correction. Parameter count: $2rH$. Crucially, this module is \textbf{position-agnostic}: the same parameters are applied identically at every token position $t$.

\subsection{Position-Dependent Bias in Decoders}
\label{sec:position-bias}

To examine whether the uniform-bias assumption of encoder-based surgery transfers to decoder models, we compute the per-token $L_1$ bias $\delta_t$ (Eq.~\ref{eq:rep-bias}) across a calibration set of prompts from each domain using Llama-2-7B merged models.

The results (Figure~\ref{fig:teaser}) reveal a striking pattern absent in encoder models: representation bias exhibits a clear \emph{increasing trend} with token position. Tokens at early positions (0--50) show relatively low bias, while tokens at later positions (100--256) exhibit significantly higher bias, often 2--3$\times$ the magnitude of early tokens. This pattern is consistent across all three domains (code, math, instruction following).

The explanation follows directly from the causal attention mechanism~\citep{vaswani2017attention}. In encoder models (e.g., ViT), bidirectional attention distributes merging errors uniformly across all tokens. In decoder models, token $t$ attends only to the preceding context $\{0, 1, \ldots, t{-}1\}$. If earlier tokens carry representation bias from merging, token $t$ aggregates these biased representations through attention, causing the bias to \emph{compound}. Each successive position accumulates not only its own merging error but also the errors of all preceding positions, producing the observed monotonically increasing pattern.

This has a direct implication for surgery module design: a position-agnostic correction, which applies the same magnitude of correction at every position, must compromise between under-correcting later tokens (where bias is largest) and over-correcting earlier tokens (where bias is smallest). This motivates the position-aware correction described below.

\subsection{Position Correction Module}
\label{sec:posdirect}

To address the position-dependent structure of representation bias, we augment the content correction with a direct per-position additive bias $b \in \R^{T_{\max} \times H}$, zero-initialized:
\begin{equation}
\hat{h}_t = h^M_t - \underbrace{c(h^M_t)}_{\text{content (input-dep.)}} - \underbrace{b_t}_{\text{position (input-indep.)}}
\label{eq:posdirect}
\end{equation}

The design decouples two sources of representation bias. The content correction $c(h^M_t)$ handles \emph{input-dependent} errors, specifically the bias that varies with the specific input at position $t$. The position bias $b_t$ captures the \emph{systematic per-position offset} that is consistent across inputs, reflecting the cumulative effect of causal error accumulation documented in Section~\ref{sec:position-bias}.

A key design choice is that $b_t$ is a \emph{free parameter} per position, rather than being parameterized by a network. Each position's correction is independent of other positions and cannot compose. This provides \emph{structural regularization} that prevents the position correction from overfitting to the small calibration set. We explored network-parameterized alternatives (viz., parallel adapters; see Appendix~\ref{sec:surgery-variants}), but found that the direct additive bias consistently matched or outperformed these more complex architectures while requiring no additional hyperparameters.

\textbf{Parameter count.} $2rH + T_{\max}H$. For Llama-2-7B ($H{=}4096$, $r{=}16$, $T_{\max}{=}512$): ${\approx}2.2\text{M}$ per domain, ${\approx}6.6\text{M}$ total for 3 domains (${\approx}0.1\%$ of 7B).

\subsection{Entropy-Weighted L1 Loss}
\label{sec:ewl1}

The standard surgery objective~\citep{yang2024representation} minimizes the $L_1$ distance between corrected and target representations, treating all token positions equally:
\begin{equation}
\mathcal{L}_{\text{L1}} = \frac{1}{H\,|\mathcal{T}|} \sum_{t \in \mathcal{T}} \|\hat{h}_t - h^*_t\|_1
\label{eq:l1-loss}
\end{equation}
where $\mathcal{T}$ is the set of non-padded positions. This uniform weighting is well-suited to classification tasks where all patch positions contribute similarly to the final prediction. In autoregressive generation, however, not all positions matter equally.

We propose weighting each position by the target model's prediction entropy. Let $p^*_{t,v} = \softmax(W_U^* h^*_t)_v$ be the target model's predicted distribution at position $t$, and let $\mathcal{H}_t = -\sum_v p^*_{t,v} \log p^*_{t,v}$ denote its entropy. The entropy-weighted $L_1$ loss is:
\begin{equation}
\mathcal{L}_{\text{EWL1}} = \frac{1}{H\,|\mathcal{T}|} \sum_{t \in \mathcal{T}} \frac{\mathcal{H}_t}{\bar{\mathcal{H}}} \|\hat{h}_t - h^*_t\|_1
\label{eq:ewl1-loss}
\end{equation}
where $\bar{\mathcal{H}} = \frac{1}{|\mathcal{T}|}\sum_{t'} \mathcal{H}_{t'}$ is the mean entropy. The unit-mean normalization ensures that the overall loss scale remains comparable to standard $L_1$, allowing the same learning rate without adjustment.

The intuition is as follows. At high-entropy positions, where the target model distributes probability across many plausible next tokens, small hidden-state errors can flip the argmax prediction, changing the token that would be generated and cascading through subsequent autoregressive steps. At low-entropy positions (e.g., deterministic syntax, common phrases, punctuation), the model's prediction is robust to perturbation: even a moderate hidden-state error leaves the argmax unchanged. By upweighting high-entropy positions, EWL1 focuses the surgery module's limited correction capacity on the positions that actually affect generation quality, analogous to how importance-weighted distillation~\citep{hinton2015distilling} focuses on informative training examples.

\section{Experimental Setup}
\label{sec:experiments}

\subsection{Models and Merging}

    We use Llama-2-7B~\citep{touvron2023llama} ($H{=}4096$, $V{=}32{,}000$) as the pretrained base and merge three publicly available fine-tunes: Llama-2-7B-Instruct (Meta), WizardMath-7B-v1.0~\citep{luo2023wizardmath}, and Llama-2-7B-Code~\citep{roziere2023code}.


We evaluate four families of merging methods across an extensive hyperparameter sweep totaling over 124 configurations. Task Arithmetic~\citep{ilharco2023editing} merges task vectors with a scaling coefficient $\lambda \in \{0.3, 0.5, 0.7, 0.9, 1.0\}$. Weight Averaging spans 22 weight distributions, including equal weighting, single-model dominant (40--80\% weight on one model), and pairwise-heavy configurations. TIES-Merging~\citep{yadav2023tiesmerging} is swept over 42 configurations combining weight $\in \{0.3, 0.5, 0.7, 1.0, 1.2, 1.5\}$ and density $\in \{0.1, 0.2, 0.3, 0.5, 0.7, 0.9\}$, including both symmetric and asymmetric per-model weight assignments. DARE~\citep{yu2024language} covers 60 configurations across both \texttt{dare\_ties} and \texttt{dare\_linear} variants with weight $\in \{0.3, 0.5, 0.7, 1.0, 1.2\}$ and density $\in \{0.1, 0.2, 0.3, 0.5, 0.7, 0.9\}$. All merged models are produced using mergekit~\citep{goddard2024arcee}. For each merging method, we select the best-performing configuration by aggregate benchmark score as the representative baseline for that method in subsequent surgery experiments.

\subsection{Evaluation Metrics}

    We evaluate on three benchmarks spanning complementary capabilities. For code generation, we use \textbf{HumanEval}~\citep{chen2021humaneval} (164 problems) and report \textbf{Pass@1}, which measures the fraction of problems solved correctly on the first attempt. For mathematical reasoning, we use the test set of \textbf{GSM8K}~\citep{cobbe2021gsm8k} (1,319 problems) with chain-of-thought zero-shot prompting~\citep{wei2022chain} and report \textbf{Accuracy}. For instruction following, we use \textbf{AlpacaEval}~\citep{dubois2024alpacaeval} (805 instructions) and report the \textbf{Win Rate} against GPT-3.5-Turbo as judged by GPT-4.1.

In addition to these task-level metrics, we report \textbf{$L_1$ Reduction} ($\rho_{{L_1}}$) as an indicator of how effectively surgery corrects representation bias in hidden-state space. Let $\delta^{\text{before}}_t = \frac{1}{H}\|h^M_t - h^*_t\|_1$ and $\delta^{\text{after}}_t = \frac{1}{H}\|\hat{h}_t - h^*_t\|_1$ denote the per-position $L_1$ residuals before and after surgery. The $L_1$ Reduction is:
\begin{equation}
\rho_{\text{$L_1$}} = 1 - \frac{\sum_{t \in \mathcal{T}} \delta^{\text{after}}_t}{\sum_{t \in \mathcal{T}} \delta^{\text{before}}_t}
\label{eq:l1-reduction}
\end{equation}
Higher $\rho_{\text{L1}}$ indicates that a larger fraction of the representation bias has been removed. As shown by~\citet{yang2024representation}, this metric correlates with downstream task performance in encoder models; we examine whether the same relationship holds for decoder models in Section~\ref{sec:analysis}.

\subsection{Implementation Details}

All merged models are produced using mergekit~\citep{goddard2024arcee} from three Llama-2-7B fine-tunes: Llama-2-7B-Instruct, WizardMath-7B-v1.0, and Llama-2-7B-Code. Surgery modules are trained for 500 iterations with the Adam optimizer ($\beta_1{=}0.9$, $\beta_2{=}0.999$, ${\eta}={10^{-3}}$) on 50 domain-specific calibration samples (HumanEval prompts for code, GSM8K questions for math, AlpacaEval instructions for instruct), tokenized to a maximum length of 512. {The calibration samples are completely excluded from the final evaluation sets to ensure that the main results represent true zero-shot performance evaluated solely on unseen test samples (see Table~\ref{tab:data-splits} for split details).} Hidden states are precomputed from both frozen models in bfloat16 and cached on GPU; only the surgery parameters are updated. For evaluation, HumanEval and GSM8K are run through lm-eval-harness~\citep{eval-harness} with greedy decoding (temperature 0.0) and the HuggingFace backend; the maximum number of generated tokens is 512 for HumanEval and 512 for GSM8K (chain-of-thought, zero-shot). AlpacaEval responses are generated with sampling (temperature 0.7, top-$p$ 0.9, max 1024 tokens) and judged pairwise against GPT-3.5-Turbo using GPT-4.1 as judge. All experiments are conducted on NVIDIA A100-SXM4-80GB GPUs.

\begin{table*}[t]
\centering
\small
\begin{tabular}{llccc|r}
\toprule
Merging Method & Surgery Method & HumanEval & GSM8K & AlpacaEval & Avg. \\
\midrule
\multirow{4}{*}{Weight Averaging}
  & None (baseline)         & 27.4 & 31.2 & 37.3 & 32.0 \\
  & \citet{yang2024representation}           & 26.2 & 28.7 & 39.5 & 31.5 \\
  & \darts\ (ours) & \textbf{26.8} & \textbf{28.9} & \textbf{41.3} & \textbf{32.3} \\
\midrule
\multirow{4}{*}{Task Arithmetic}
  & None (baseline)         & 23.8 & 41.2 & 49.5 & 38.2 \\
  & \citet{yang2024representation}           & 23.2 & 42.5 & 47.0 & 37.6 \\
  & \darts\ (ours) & \textbf{25.0} & \textbf{43.4} & \textbf{58.6} & \textbf{42.3} \\
\midrule
\multirow{4}{*}{TIES-Merging}
  & None (baseline)         & 17.1 & 12.6 & 8.9 & 12.9 \\
  & \citet{yang2024representation}           & \textbf{23.2} & 7.3 & 12.8 & \textbf{14.4} \\
  & \darts\ (ours) & {21.9} & \textbf{8.3} & \textbf{13.0} & \textbf{14.4} \\
\midrule
\multirow{4}{*}{DARE}
  & None (baseline)         & 19.5 & 34.9 & 47.7 & 34.0 \\
  & \citet{yang2024representation}           & 18.3 & \textbf{35.4} & 47.6 & 33.8 \\
  & \darts\ (ours) & \textbf{21.3} & {34.9} & \textbf{48.7} & \textbf{35.0} \\
\bottomrule
\end{tabular}
\caption{Main results on three benchmarks. Comparing the original surgery approach (Standard + $L_1$) against our decoder-specific adaptations. Best results per merged model in \textbf{bold}.}
\label{tab:main-results}
\end{table*}

\subsection{Baselines}

We compare four settings. \textbf{Individual models}: each fine-tuned model evaluated on all three benchmarks, establishing the upper bound for domain-specific performance. \textbf{Merged (no surgery)}: the merged model evaluated directly. \textbf{Standard + $L_1$}~\citep{yang2024representation}: position-agnostic surgery with standard $L_1$ loss, representing the original approach. \textbf{\darts} (ours): position correction module with entropy-weighted $L_1$ loss.

\begin{table}[t]
\centering
\small
\begin{tabular}{lrrr}
\toprule
Model & HE & GSM & AE \\
\midrule
Llama-2-7B-Base & 12.8 & 4.5 & 1.5 \\
Llama-2-7B-Instruct & 15.2 & 24.5 & 58.2 \\
WizardMath-7B-v1.0 & 12.8 & 39.7 & 7.0 \\
Llama-2-7B-Code & 28.1 & 5.2 & 0.2 \\
\bottomrule
\end{tabular}
\caption{Individual model baselines. HE = HumanEval (pass@1), GSM = GSM8K (exact match), AE = AlpacaEval (win rate \%).}
\label{tab:individual-baselines}
\end{table}

\section{Results}
\label{sec:results}

\subsection{Main Results}

Table~\ref{tab:main-results} presents the main results, comparing the original surgery approach (Standard + $L_1$) against \darts\ across both merged model configurations and all three benchmarks.

\section{Analysis}
\label{sec:analysis}

\subsection{L1 Bias Reduction vs Task Performance}

A central question is whether reducing representation bias in hidden-state space translates to downstream task performance. Table~\ref{tab:bias-vs-perf} reports the mean $L_1$ bias before and after \darts\ surgery alongside the corresponding task performance changes.

\begin{table*}[t]
\centering
\small
\begin{tabular}{llrrrrr}
\toprule
Merged Model & Domain & $\rho_{{L}_1}$ & \multicolumn{2}{c}{Task Score} & pp \\
\cmidrule(lr){4-5}
             &        & (\%) & No Surg. & \darts & Change \\
\midrule
\multirow{3}{*}{TA ($\lambda = 0.5$)} & Code     & 8.2 & 23.8 & 25.0 & +1.2 \\
                                   & Math     & 11.3 & 41.2 & 43.4 & +2.2 \\
                                   & Instruct & 42.9 & 49.5 & 58.6 & +9.1 \\
\midrule
\multirow{3}{*}{TA ($\lambda = 0.7$)} & Code     & 14.2 & 21.9 & 26.8 & +4.9 \\
                                   & Math     & 14.3 & 46.2 & 42.9 & $-$3.3 \\
                                   & Instruct & 51.0 & 49.3 & 59.2 & +9.9 \\
\bottomrule
\end{tabular}
\caption{Correlation between $L_1$ bias reduction and task performance. pp = percentage point change in the task metric.}
\label{tab:bias-vs-perf}
\end{table*}

\subsection{Statistically Significant Improvements}

To verify that the improvements of \darts\ over the standard surgery baseline are not due to random variation in calibration data selection, we train both methods across 6 different random seeds (controlling the calibration data shuffle order) and compare their average scores. Table~\ref{tab:stat-sig} reports the per-seed average performance (mean of HumanEval, GSM8K, and AlpacaEval) for each method.

We first test each group for normality using the Shapiro-Wilk test ($\alpha = 0.05$). The \darts\ scores pass the normality test ($W = 0.80$, $p = 0.059$), while the Standard surgery scores do not ($W = 0.77$, $p = 0.034$). Since one group is non-normal, we apply the Mann-Whitney U test rather than a parametric $t$-test. The result ($U = 36.0$, $p = 0.002$) indicates that \darts\ significantly outperforms Standard surgery at the $\alpha = 0.05$ level. Despite a modest absolute improvement of 2.0 percentage points in average score, the consistency of the gain across all 6 seeds (with \darts\ outperforming Standard in every run) yields a highly significant $p$-value, confirming that the improvement is robust to calibration data variation. {A more granular statistical analysis is given in \appendixname~\ref{sec:granular-stats}}

\begin{table}[t]
\centering
\small
\begin{tabular}{lrr}
\toprule
Seed & \darts\ (Avg.) & Standard (Avg.) \\
\midrule
default & 42.3 & 37.6 \\
7       & 42.2 & 41.7 \\
29      & 42.3 & 40.4 \\
71      & 43.3 & 41.8 \\
164     & 42.2 & 41.2 \\
190     & 43.0 & 41.1 \\
\midrule
Mean $\pm$ Std & 42.6 $\pm$ 0.5 & 40.6 $\pm$ 1.6 \\
\midrule
\multicolumn{3}{l}{\textit{Mann-Whitney U test}: $U = 36.0$, $p = 0.002$} \\
\bottomrule
\end{tabular}
\caption{Statistical significance of \darts\ vs.\ Standard surgery ($L_1$ loss). Average score = mean of HumanEval, GSM8K, and AlpacaEval across 6 seeds.}
\label{tab:stat-sig}
\end{table}


\section{Ablation Studies}
\label{sec:ablations}

\subsection{Loss Function Ablation}

 Table~\ref{tab:loss-ablation} compares four loss functions using Standard (position-agnostic) surgery on TA $(\lambda = 0.5)$, holding architecture constant to isolate the effect of the loss. From the results, we find that \emph{loss function choice matters more than surgery module architecture} for decoder-based surgery.

\begin{table}[t]
\centering
\small
\begin{tabular}{lrrr|r}
\toprule
Loss Function & HE & GSM & AE & Avg. \\
\midrule
$L_1$ (original)           & 23.2 & 42.5 & 47.0 & 37.6 \\
KL on logits            & 20.1 & \textbf{43.4} & 54.5 & 39.3 \\
Cross-entropy distill.  & 18.3 & 33.1 & 40.0 & 30.5 \\
$\text{EWL}_1$ (ours) & \textbf{25.6} & {41.7} & \textbf{{57.0}} & \textbf{41.4} \\
\bottomrule
\end{tabular}
\caption{Loss function ablation with Standard surgery. EWL1 dominates, confirming that loss choice is the primary driver of improvement.}
\label{tab:loss-ablation}
\end{table}

\subsection{Surgery Module Ablation}

Table~\ref{tab:module-ablation} compares Standard surgery against \darts\ (with position correction), both using EWL1 loss, isolating the complementary contribution of the position correction module. From the results, we find that adding the position correction module leads to modest but significant improvements.

\begin{table}[t]
\centering
\small
\begin{tabular}{lrrr}
\toprule
Module & HE & GSM & AE \\
\midrule
Standard                & 25.6 & 41.7 & 57.0 \\
+ Position corr.\ (ours) & \textbf{25.0} & \textbf{43.4} & \textbf{58.6} \\
\midrule
$\Delta$ & $-$0.6 & +1.7 & +1.6 \\
\bottomrule
\end{tabular}
\caption{Surgery module ablation with EWL1 loss. Position correction provides complementary gains on top of the improved loss.}
\label{tab:module-ablation}
\end{table}

\subsection{Learning Rate Sensitivity}

\begin{figure}[t]
  \centering
  \includegraphics[width=\columnwidth]{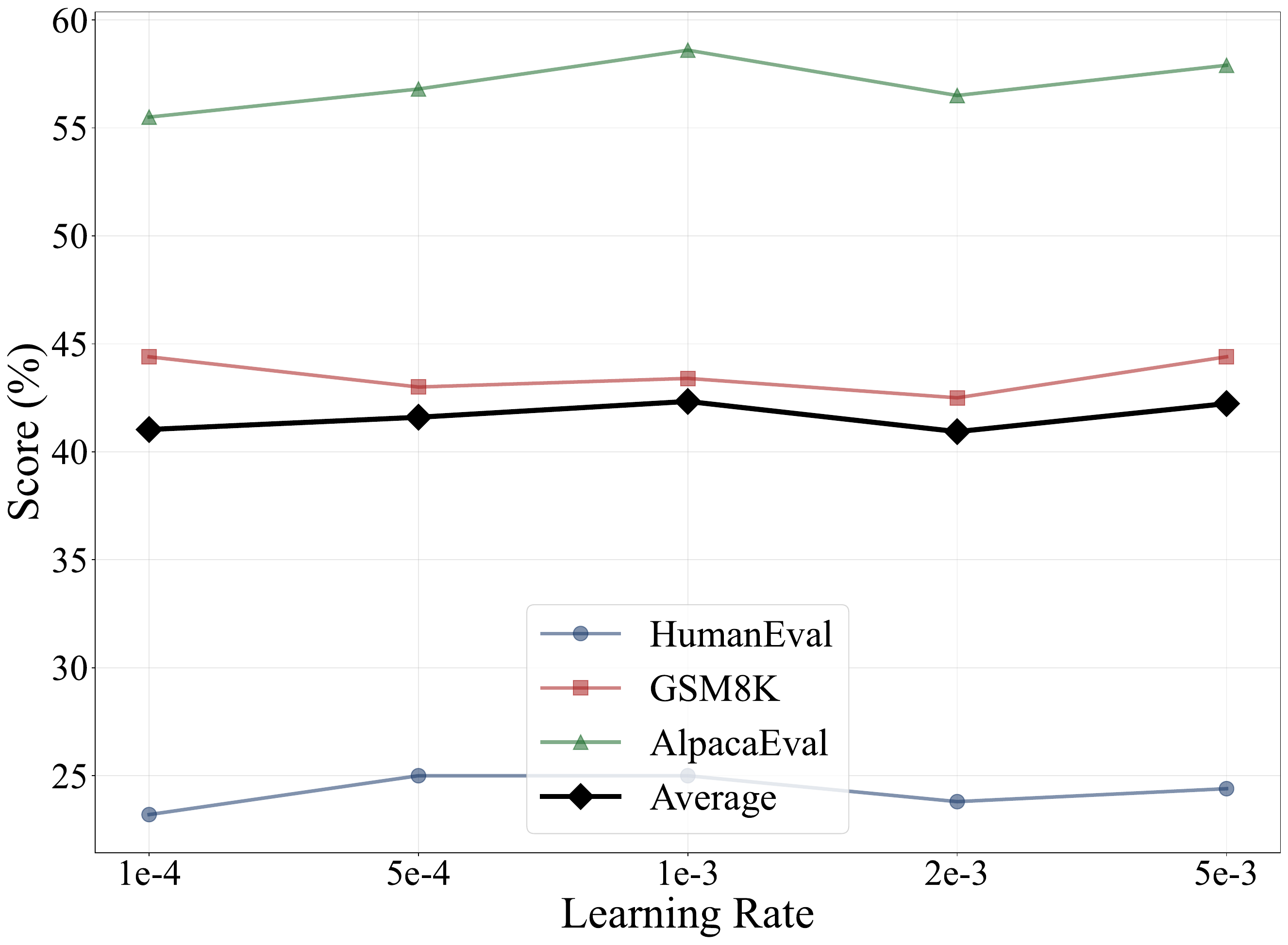}
  \caption{Learning rate sensitivity of \darts. Performance is stable across two orders of magnitude, with the default $10^{-3}$ achieving the best average.}
  \label{fig:lr-sensitivity}
\end{figure}

Figure~\ref{fig:lr-sensitivity} shows \darts\ performance across learning rates spanning two orders of magnitude ($10^{-4}$ to $5 \times 10^{-3}$). Performance is remarkably stable: the average score varies by less than 2 percentage points across the entire range, and no single learning rate causes catastrophic degradation on any benchmark. The default learning rate of $10^{-3}$ achieves the best AlpacaEval score and the highest average, but nearby values ($5 \times 10^{-4}$, $5 \times 10^{-3}$) perform competitively. This robustness suggests that \darts\ does not require careful learning rate tuning in practice.

\section{Conclusion}
\label{sec:conclusion}

We have extended representation surgery from encoder-based vision models to decoder-based language models and identified two fundamental challenges specific to the autoregressive setting: position-dependent bias accumulation under the causal attention mask, and non-uniform position importance arising from the structure of language generation. To address these challenges, we proposed \darts, which combines entropy-weighted $L_1$ loss (focusing correction on decision-critical positions) with a structurally regularized position correction module that captures the monotonically increasing bias pattern. Our experiments on Llama-2-7B merged models across three diverse domains demonstrate that {loss function design is the dominant factor}: entropy-weighted $L_1$ alone accounts for the majority of performance gains, with position correction providing complementary improvements. On Llama-2-7B merged models across code, math, and instruction-following domains, \darts\ achieves significant improvement over the original surgery approach.

\section*{Limitations}

\begin{enumerate}
    \item \textbf{Sequence length generalization.} The position correction bias table $b \in \R^{T_{\max} \times H}$ cannot extrapolate beyond $T_{\max} = 512$. Positions exceeding $T_{\max}$ receive zero position correction (content correction still applies). Future work could explore continuous position functions or relative position encodings.

    \item \textbf{Scale.} We evaluate exclusively on Llama-2-7B. Whether \darts\ transfers to larger models (13B, 70B) or different architectures (Mistral~\citep{jiang2023mistral}) remains untested.

    \item \textbf{Domain asymmetry.} The entropy weighting may deprioritize positions critical for code generation (e.g., syntactically deterministic but semantically critical tokens), potentially explaining domain-dependent gains.
\end{enumerate}


\bibliography{custom}

\appendix

\section{{Calibration vs. Test split}}
\begin{table}[h]

\centering
\scriptsize
\setlength{\tabcolsep}{2pt}
\begin{tabular}{lccc}
\toprule
\textbf{Benchmark} & \textbf{Total Samples} & \textbf{Calibration Split} & \textbf{Evaluation Split} \\
\midrule
HumanEval (Code) & 164 & 50 & 114 \\
GSM8K (Math) & 1,319 & 50 & 1,269 \\
AlpacaEval (Instruct) & 805 & 50 & 755 \\
\bottomrule
\end{tabular}
\caption{{Dataset partitions detailing the calibration (training) and evaluation (test) splits for each benchmark. The splits are mutually exclusive to prevent any potential leakage.}}
\label{tab:data-splits}
\end{table}

\section{{Calibration Cost}}
\label{sec:calib-cost}

{Although \darts\ introduces negligible inference overhead ($\approx$0.1\% additional parameters), the offline calibration stage requires forward passes through both the merged and target models to precompute hidden states and prediction entropy weights. To make the practical resource requirements fully transparent, we measure the end-to-end calibration cost on a single NVIDIA A100-SXM4-80GB GPU using Llama-2-7B-Code as the target model, with the standard protocol: 50 unlabeled calibration prompts, $T_{\max} = 512$, batch size 8, 500 training iterations, rank-16 surgery bottleneck.}

{Table~\ref{tab:calib-cost} reports the per-configuration cost (one merged model $\times$ one domain $\times$ one surgery variant). The actual surgery computation (precompute + training) takes only $\approx$4--5 seconds; loading the two 7B models accounts for $\approx$7 seconds and dominates the wall-clock time. Peak GPU memory remains well within A100 capacity at $\approx$30--31.5 GB (under 40\% utilization). The hidden-state cache occupies 0.42 GB, and the trained surgery module is 265 KB--4.46 MB on disk ($<$0.04\% of backbone parameters). Since each domain is calibrated independently, the total cost for all three domains is $\approx$36 seconds, making the entire calibration pipeline lightweight enough to run on a single commodity GPU.}

\begin{table}[h]

\centering
\small
\setlength{\tabcolsep}{2pt}
\begin{tabular}{lrr}
\toprule
\textbf{Phase} & \textbf{\citet{yang2024representation}} & \textbf{\darts} \\
\midrule
Load merged model & 3.37\,s & 3.46\,s \\
Load target model & 3.43\,s & 3.37\,s \\
Precompute (2$\times$ forward) & 1.81\,s & 1.87\,s \\
Training loop (500 iter) & 2.40\,s & 3.15\,s \\
\midrule
Total wall-clock & 11.01\,s & 11.85\,s \\
Peak GPU memory & 29.91\,GB & 31.49\,GB \\
Hidden-state cache & 0.42\,GB & 0.42\,GB \\
Surgery parameters & 131,072 & 2,228,224 \\
Checkpoint size on disk & 265\,KB & 4.46\,MB \\
\bottomrule
\end{tabular}
\caption{{Per-domain calibration cost on a single NVIDIA A100 (80\,GB). Model: Llama-2-7B-Code ($\approx$7B params), 50 calibration prompts, $T_{\max} = 512$, rank 16, 500 iterations. The surgery computation (precompute + training) is $\approx$4--5\,s; model loading ($\approx$7\,s) dominates.}}
\label{tab:calib-cost}
\end{table}

\section{Effect of Calibration Data Size}
\label{sec:calib-size}

The surgery module is trained on a small set of $N$ domain-specific calibration samples. Table~\ref{tab:calib-size} examines how performance varies as $N$ ranges from 10 to the full test set. Performance is relatively stable across calibration sizes, with as few as 20 samples achieving results competitive with larger budgets. This suggests that the surgery module learns a generalizable bias correction rather than memorizing calibration-specific patterns. Using the full test set for calibration yields comparable results, confirming that the module does not overfit even with abundant data. {Note that calibrating on the full test set is strictly an exploratory analysis intended to examine data scaling limits and establish a theoretical upper bound; these test-adapted results are kept isolated to this appendix and are not used for any of the main model claims or baseline comparisons in the paper.}

\begin{table}[h]
\centering
\small
\begin{tabular}{rrrr}
\toprule
Calibration $N$ & HE & GSM & AE \\
\midrule
10  & 24.4 & 42.5 & 54.0 \\
20  & 24.4 & 44.8 & 58.4 \\
30  & 26.2 & 43.8 & 56.8 \\
50  & 25.0 & 43.4 & 58.6 \\
100 & 26.2 & 42.8 & 55.4 \\
Full Test & 23.17 & 43.52 & 59.45 \\
\bottomrule
\end{tabular}
\caption{Effect of calibration data size.}
\label{tab:calib-size}
\end{table}

\section{Full Loss Function Comparison}
\label{sec:full-loss-grid}

In the main paper (Table~\ref{tab:loss-ablation}), we compare loss functions using Standard surgery only. Here, Table~\ref{tab:full-loss-grid} presents the full factorial comparison of loss functions crossed with surgery module type (Standard vs.\ position correction). This allows us to disentangle the contribution of the loss function from the contribution of the position correction module. Two patterns emerge. First, switching from $L_1$ to entropy-weighted $L_1$ consistently improves average performance regardless of the surgery module, confirming that the loss function is the primary driver. Second, adding position correction on top of each loss function provides a further boost in average score, with the largest gain observed for EWL1 (+0.9 average), indicating that the two innovations are complementary.

\begin{table*}[h]
\centering
\small
\begin{tabular}{llccc|r}
\toprule
Loss Function & Surgery Module & HumanEval & GSM8K & AlpacaEval & Avg. \\
\midrule
$L_1$                   & Standard  & 23.2 & 42.5 & 47.0 & 37.6 \\
$L_1$                   & + Pos.\ corr. & 21.3 & 43.1 & 48.8 & 37.7 \\
Entropy-Weighted $L_1$  & Standard  & 25.6 & 41.7 & 57.0 & 41.4 \\
Entropy-Weighted $L_1$  & + Pos.\ corr. & 25.0 & 43.4 & 58.6 & 42.3 \\
\bottomrule
\end{tabular}
\caption{Full loss function $\times$ surgery module grid.}
\label{tab:full-loss-grid}
\end{table*}

\section{Position Correction Module Variants}
\label{sec:surgery-variants}

Beyond the position correction module used in \darts, we explored several alternative architectures for incorporating position information into the surgery module. Table~\ref{tab:surgery-variants} compares these variants, all trained with EWL1 loss at rank 16. \textbf{Standard} applies only the content bottleneck with no position awareness. \textbf{Add} adds a learned position embedding to the hidden state before the down-projection, allowing position to modulate the ReLU activation pattern. \textbf{Parallel} uses two independent bottleneck paths (one for content, one for position) whose outputs are summed; however, we observe that the position path causes severe degradation at inference, likely due to overfitting. \textbf{Pos.\ Corr.\ (Ours)} adds a free per-position bias vector that is independent across positions and zero-initialized, providing structural regularization that prevents overfitting while still capturing systematic position-dependent offsets. Among all variants, our position correction module achieves the highest average score while maintaining stable HumanEval performance, unlike the Parallel variant which collapses on code generation.

\begin{table}[h]
\centering
\small
\resizebox{\linewidth}{!}{

\begin{tabular}{lrrrr}
\toprule
Variant & HE & GSM & AE & Extra Params \\
\midrule
Standard    & 25.6 & 41.7 & 57.0 & 0 \\
Add    & 21.9 & 44.2 & 57.9 & 2.1M \\ 
Parallel & 12.2 & 45.0 & 55.5 & 2.2M \\ 
Pos. Corr. (Ours)   & 25.0 & 43.4 & 58.6 & 2.1M \\
\bottomrule
\end{tabular}
}
\caption{Position correction module variant comparison.}
\label{tab:surgery-variants}
\end{table}

\section{Effect of Training Steps}
\label{sec:steps-ablation}

Table~\ref{tab:steps-ablation} varies the number of training iterations from 50 to 1000. Performance is stable across this range, with even 50 iterations producing competitive results. Longer training (1000 steps) yields a modest GSM8K improvement without degrading other benchmarks.

\begin{table}[h]
\centering
\small
\begin{tabular}{rrrr}
\toprule
Steps & HE & GSM & AE \\
\midrule
50   & 25.6 & 42.3 & 57.9 \\
100  & 25.0 & 42.9 & 56.5 \\
200  & 25.0 & 43.4 & 57.9 \\
400  & 24.4 & 43.0 & 57.0 \\
500 (default) & 25.0 & 43.4 & 58.6 \\
1000 & 25.6 & 44.3 & 58.4 \\
\bottomrule
\end{tabular}
\caption{Effect of training iterations.}
\label{tab:steps-ablation}
\end{table}

\section{Effect of Surgery Rank}
\label{sec:rank-ablation}

Table~\ref{tab:rank-ablation} varies the bottleneck rank $r$ of the surgery module. Rank 16 (the default) achieves the best overall balance. Increasing rank to 32 or 64 does not improve performance and slightly hurts AlpacaEval, while rank 128 degrades HumanEval, suggesting that higher-rank modules overfit on the small calibration set.

\begin{table}[t]
\centering
\small
\begin{tabular}{rrrrl}
\toprule
Rank $r$ & HE & GSM & AE & Params/domain \\
\midrule
16  & 25.0 & 43.4 & 58.6 & 131K + 2.1M \\
32  & 25.0 & 44.3 & 56.4 & 262K + 2.1M \\
64  & 25.0 & 42.8 & 56.0 & 524K + 2.1M \\
128  & 23.8 & 42.6 & 58.1 & 1.04M + 2.1M \\
\bottomrule
\end{tabular}
\caption{Effect of surgery rank on \darts.}
\label{tab:rank-ablation}
\end{table}

\section{Seed Sensitivity}
\label{sec:seed-sensitivity}

Table~\ref{tab:seed-sensitivity} reports \darts\ performance across 6 random seeds that control the calibration data shuffle order. The low standard deviation (HE: $\pm$0.2, GSM: $\pm$0.8, AE: $\pm$1.0) confirms that \darts\ is robust to the choice of calibration samples.

\begin{table}[h]
\centering
\small
\begin{tabular}{rrrr}
\toprule
Seed & HE & GSM & AE \\
\midrule
default & 25.0 & 43.4 & 58.6 \\
7       & 24.4 & 45.0 & 57.3 \\
29      & 25.0 & 43.6 & 58.3 \\
71      & 25.0 & 45.0 & 59.9 \\
164     & 25.0 & 44.4 & 57.1 \\
190     & 25.0 & 45.0 & 59.0 \\
\midrule
Mean $\pm$ Std & 24.9 $\pm$ 0.2 & 44.4 $\pm$ 0.8 & 58.4 $\pm$ 1.0 \\
\bottomrule
\end{tabular}
\caption{Reproducibility across random seeds. Seed controls the calibration data shuffle order.}
\label{tab:seed-sensitivity}
\end{table}

\section{{Granular Statistical Analysis of Seed Variation}}
\label{sec:granular-stats}

{To provide a more granular statistical analysis, we compare the task-level performance of \darts\ against the Standard surgery baseline across the 6 varied calibration seeds (including the default seed) using both parametric and non-parametric tests. We perform paired $t$-tests and Wilcoxon signed-rank tests on the raw scores for each task. The results, summarized in Table~\ref{tab:granular-stats}, show that \darts\ provides statistically significant improvements on GSM8K and AlpacaEval using the non-parametric Wilcoxon test, and significant to marginally significant improvements using the parametric $t$-test. Performance on HumanEval remains statistically indistinguishable from the baseline.}

{Furthermore, we fit a linear mixed-effects model (\texttt{Score $\sim$ Model + Task}) with \texttt{Seed} as a random intercept across all tasks. The coefficient for the \darts\ model is positive and highly significant ($\beta = 0.019$, $z = 3.224$, $p = 0.001$), confirming that \darts\ provides a consistent and statistically significant improvement over the baseline across tasks and seeds.}

\begin{table}[h]
\centering
\scriptsize
\setlength{\tabcolsep}{2pt}
\begin{tabular}{lrrrrr}
\toprule
\textbf{Task} & \textbf{Mean Diff} & \textbf{$t$-stat} & \textbf{$t$-test $p$-val} & \textbf{$W$-stat} & \textbf{Wilcoxon $p$-val} \\
\midrule
HumanEval & +0.30 & 0.690 & 0.521 & 7.500 & 0.688 \\
GSM8K & +1.56 & 2.849 & 0.036 & 0.000 & 0.031 \\
AlpacaEval & +3.95 & 2.459 & 0.057 & 0.000 & 0.031 \\
\bottomrule
\end{tabular}
\caption{{Paired $t$-test and Wilcoxon signed-rank test comparing \darts\ vs.\ Standard surgery across 6 random seeds for each task. Positive Mean Diff indicates \darts\ outperforms the baseline.}}
\label{tab:granular-stats}
\end{table}

\end{document}